%% file: acl.tex
\pdfoutput=1

\documentclass[11pt]{article}

\usepackage[]{acl}

\usepackage{times}
\usepackage{latexsym}
\usepackage{soul}
\usepackage{url}
\usepackage[utf8]{inputenc}
\usepackage{graphicx}
\usepackage{amsmath}
\usepackage{amsthm}
\usepackage{booktabs}
\usepackage{algorithm}
\usepackage{algorithmic}
\usepackage{array}
\usepackage{color, colortbl}
\usepackage{multirow}
\usepackage{setspace}
\usepackage{autobreak}
\usepackage{amsfonts}
\usepackage{amssymb}
\usepackage{amsthm,amsmath}
\usepackage{mathrsfs}
\usepackage{multirow}
\usepackage{subfigure}
\usepackage{makecell}
\usepackage{pifont}
\usepackage{fontawesome}
\usepackage{bm}

\usepackage{hyperref}

\usepackage[T1]{fontenc}

\usepackage[utf8]{inputenc}

\usepackage{microtype}

\title{CausalAPM: Generalizable Literal Disentanglement for NLU Debiasing}

\author{
    { Songyang Gao$^{1*}$, \ \ Shihan Dou$^{1}$\thanks{{ }{ } Equal contribution.} , \ \ Junjie Shan$^{2}$,} \ \
    { \bf Qi Zhang$^{13}$, \ \ Xuanjing Huang$^{1}$}\thanks{{ }{ }{ }Corresponding author.} \\
    {$^1$  School of Computer Science, Fudan University, Shanghai, China} \\
    {$^2$  KTH Royal Institute of Technology, Stockholm, Sweden} \\
    {$^3$  Shanghai Key Laboratory of Intelligent Information Processing, Fudan University} \\
    \texttt{ \{gaosy21,shdou21\}@m.fudan.edu.cn}\\
}

\begin{document}

\maketitle

\input{outline/abstract}

\input{outline/intro}

\input{outline/motivation}
\input{outline/background}

\input{outline/method}
\input{outline/exp}

\input{outline/relatedwork}
\input{outline/conclusion}

\bibliography{anthology,custom}
\bibliographystyle{acl_natbib}

\end{document}

%% file: outline/abstract.tex
\begin{abstract}


Dataset bias, i.e., the over-reliance on dataset-specific literal heuristics, is getting increasing attention for its detrimental effect on the generalization ability of NLU models. Existing works focus on eliminating dataset bias by down-weighting problematic data in the training process, which induce the omission of valid feature information while mitigating bias. In this work, We analyze the causes of dataset bias from the perspective of causal inference and propose CausalAPM, a generalizable literal disentangling framework to ameliorate the bias problem from feature granularity. The proposed approach projects literal and semantic information into independent feature subspaces, and constrains the involvement of literal information in subsequent predictions.
Extensive experiments on three NLP benchmarks (MNLI, FEVER, and QQP) demonstrate that our proposed framework significantly improves the OOD generalization performance while maintaining ID performance.

\end{abstract}

%% file: outline/intro.tex
\section{Introduction}
Natural Language Understanding (NLU) aims to train machines on comprehension of structure and meaning of human language.
Pre-trained language models, like BERT, have achieved remarkable performance on NLU benchmarks \cite{wang2018glue}. 
However, recent observations \cite{mccoy2019right,naik2018stress} show that, NLU models tend to over-rely on specific shallow heuristics instead of capturing underlying semantics, resulting in inadequate generalization capability in out-of-distribution (OOD) settings \cite{schuster-etal-2019-towards}. 
In addition, \citet{sinha2020unnatural, pham2020out} have reported the insensitivity to word-order permutations among transformer-based models. When permuted randomly, both the original example and the out-of-order one elicit the same classification label, which is contradict to the conventional understanding of semantics. These phenomena are referred to as dataset bias problems.


\begin{figure}[htbp]
\centerline{\includegraphics[width=0.45\textwidth]{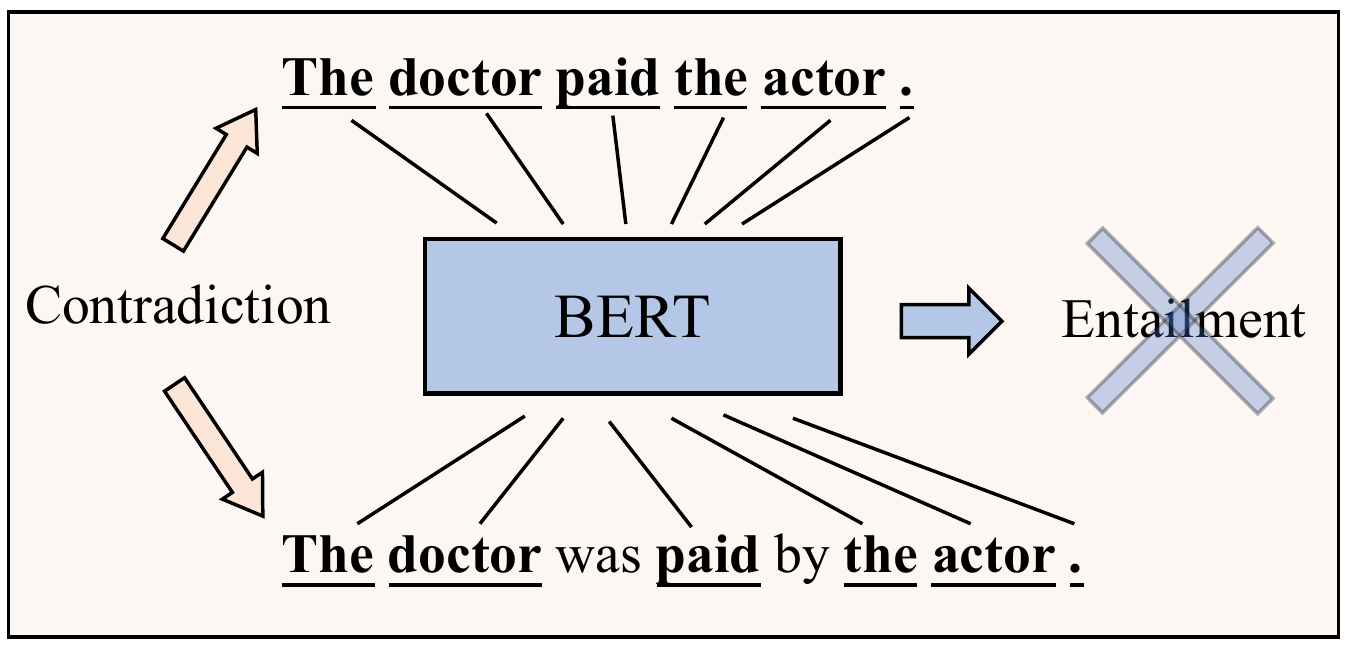}}
\caption{An example indicating dataset bias. "The doctor paid the actor" is contradict to "The doctor was paid by the actor". However, with almost identical words employed in the two sentences, BERT predicts "Entailment" for the above sentence pair.}
\label{fig:1}
\vspace{-0.5em}
\end{figure}

Existing works tend to eliminate dataset bias by reducing the negative impact of problematic data. One strategy is identifying or constructing counterexamples to existing biases, and then focus the main model on those hard minorities, such as learned-mixin \cite{clark2020learning}, example reweighting \cite{schuster-etal-2019-towards}, or confidence regularization \cite{utama2020mind}.
The other strategy depends on the specific assumption that dataset biases can be known as a prior with limited capacity models \cite{utama2020towards, sanh2020learning} or early training \cite{tu2020empirical}. 
However, these methods are not end-to-end, accompanied by a complicated training process. Furthermore, weak-weighted bias samples at data granularity simultaneously obstruct learning from their non-bias parts, resulting in a drop on the in-distribution (ID) datasets \cite{wen2021debiased}.

\begin{figure*}[htbp]
\centerline{\includegraphics[width=0.86\textwidth]{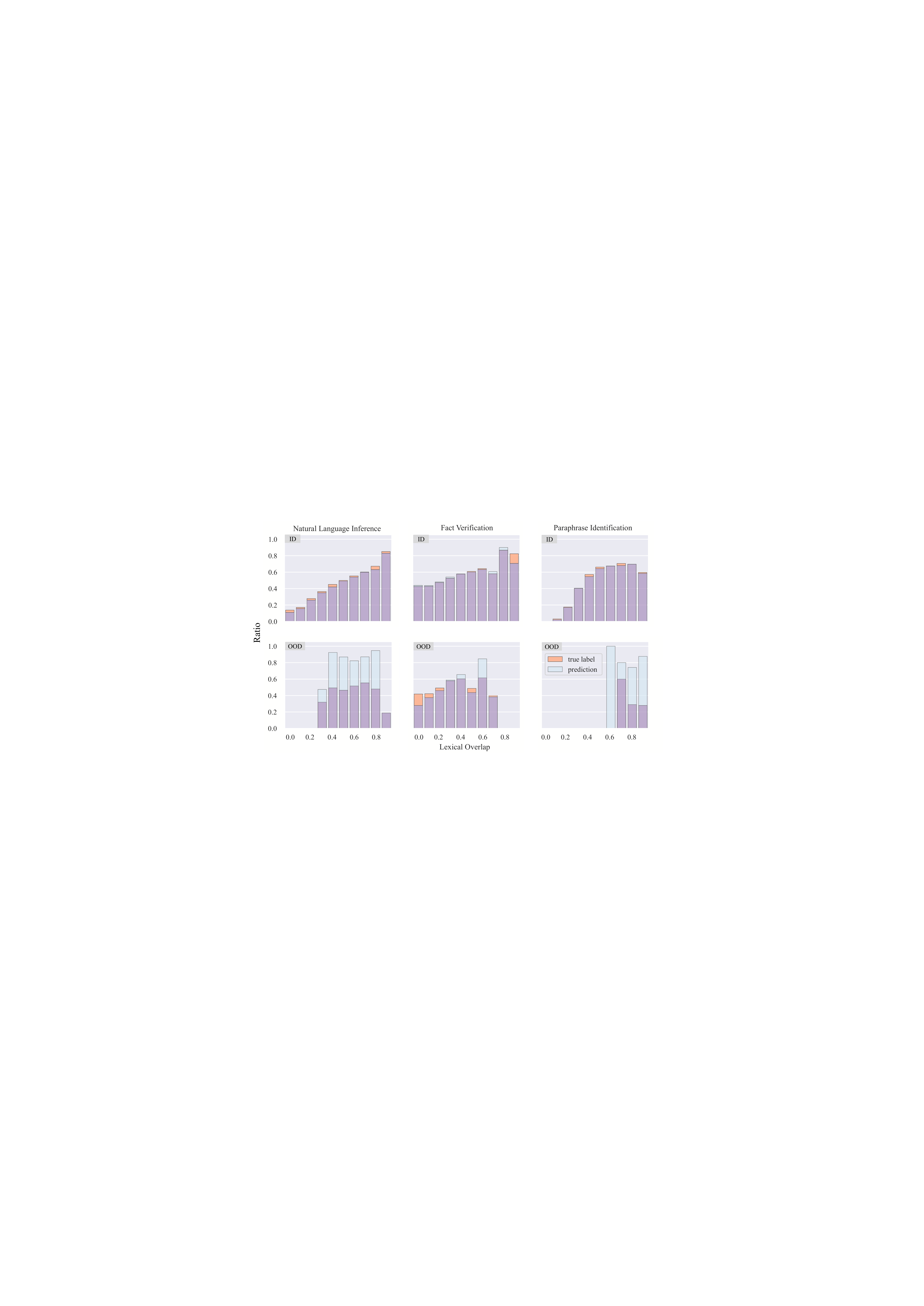}}
\caption{Predict tendency with increasing lexical overlap for syntactic label. A Bert-base-uncased model is fine-tuned on ID dataset, and evaluated on both ID and OOD dataset. Model's proportions for predict label are practical unanimity to gold label on the ID dataset, but deviating under generalization settings, which verifies the impairment of model's generalization ability by dataset bias. }
\label{fig:2}
\vspace{-0.5em}
\end{figure*}

The abovementioned trade on ID and OOD tasks inspires us to study debiasing from a fine-grained perspective. Motivated by Predictability Minimization (PM) \cite{schmidhuber2020generative}, we propose a novel learning framework-Causal Adversarial Predictability Minimization (CausalAPM). The proposed method trains an encoder to extract and weaken literal bias while maintaining semantic information by generalizable disentangled representation learning (DRL).
Specifically, CausalAPM contains two adversarial learning objectives: i) Literal information maximization, which aims to maximize the heuristic-related information extracted from sentence representations; ii) Dependence minimization, which prevents the model from separating excessive features, causing detriment to semantic information. Overall, Our main contributions are summarized as follows:
\begin{itemize}
\item We analyze multiple existing generalization tasks to verify the wild existence of literal heuristic and propose a Structural Causal Model (SCM) to model this generalization hinder during fine-tune. 
\item We evaluate the disentanglement performance of the VAE-based model on generalization tasks, demonstrate the necessity for a more generalizable disentangle model.
\item We propose CausalAPM, a causal-based adversarial disentangle framework, Extensive experiments validate the competitive effectiveness of our approach for overcoming literal heuristics while maintaining in-distribution performances.
\end{itemize}

%% file: outline/motivation.tex
\section{Motivation}

In this section, We present two preliminary experiments.
We verify the universal existence of literal heuristics in discovered bias datasets, and observe that existing VAE-based disentangle methods underperform on aforementioned generalizing tasks. We, therefore, propose CausalAPM, an adversarial disentangling framework, which constrains literal inductive biases to achieve constant generalization performance. We demonstrate how we extend PM to the debias task and address these issues in section 4.

\subsection{Literal Heuristics}

\label{sec:2}
We fine-tune the Bert-base-uncased model on MNLI, FEVER, and QQP datasets and additionally test their performance on HANS, SYMM, and PAWS datasets. Figure \ref{fig:1} verifies the heuristic captured by model during the training process. As positive samples increase with high lexical overlap, the model tends to predict specific label for high overlap instances on OOD datasets, e.g., "Entailment" for HANS. While MNLI and QQP are constructed with higher overlap bias, FEVER is slightly more gentle with such defects, however, a positive correlation between the predicted label and overlap severity can still be observed. Overall, the over-reliance on literal heuristics is a universal detriment to the model's ability to generalize.

\subsection{Debiasing with $\bm{\beta}$-VAE}
Previous works have proposed that extracted disentangled representations can improve generalization and robustness across downstream tasks \cite{higgins2016beta, bengio2013representation}. We test the $\beta$-VAE disentangle method with consistent settings to current debiasing models on three NLU tasks with eight datasets. Table \ref{tab:r1} shows the improvement in generalization by disentangling. The VAE-based method exhibit superior results to original models. However, the results on the OOD dataset are weaker relative to prior debiasing works. We argue that unsupervised disentanglement has indeed separated generative factors in the data representation, but failed on eliminate the abovementioned literal heuristics caused by unbalanced label distribution in datasets. Besides, while separated factors are independent of each other, they may consist of a combination of literal and semantic information, which induced a weaker bias.


%% file: outline/background.tex
\section{Background}
In this section, we highlight the predictability minimization principle. Subsequently, the analysis of possible issues when applying it to literal disentanglement was provided.

\subsection*{Predictability Minimization (PM)} PM principle originated in unsupervised minimax game. It attempts to achieve a disentangled factorial code of given data without assumptions to prior distribution of input data.
The code components are statistically independent of each other, which facilitates subsequent downstream learning.

Given an input data $(X,Y)$, autoencoder try to learn a reasonable low-dimensional embedding $P(Z_1,...,Z_n|X)$ to reconstruct $X$, where $\{Z_1,...,Z_n\}$ is the hidden representation of input data. Considering a subset of the feature vector M = $\{Z_1,...,Z_k|k<n\}$, PM eliminates the correlation between $M$ and it's complementary set $M^Z$ by empirical estimating the distribution of $P(M|M^Z)$ and $ P(M)$, which is equivalent to minimize the conditional entropy:
\begin{equation}
\begin{aligned}
H(M|M^Z) =  -\int_Z {P(Z)}log(P(M|M^Z))
\end{aligned}
\end{equation}


\begin{table}[ht] 
  \centering
  \caption{$\beta$-VAE performance on MNLI, Fever, QQP, and their respective challenge test sets.}
\begin{tabular}{|l|cc|}
\toprule
\multicolumn{1}{|c|}{\textbf{Dataset}} & \textbf{bert-base} & \textbf{$\bm{\beta}$-VAE}  \\
\midrule
{MNLI} & 84.3  & 84.7 \\
{HANS} & 61.1  & 65.6 \\
\midrule
{FEVER} & 85.4  & 85.5 \\
{Symm. v1} & 55.2  & 58 \\
{Symm. v2} & 63.1  & 64.8 \\
\midrule
{QQP} & 91    & 90.7 \\
{PAWS dupl} & 96.9  & 81 \\
{PAWS $\neg$ dupl} & 9.8   & 24 \\
\bottomrule
\end{tabular}%
\vspace{-0.5em}
\label{tab:r1}
\end{table}

In this way, disentangled factors are achieved which satisfying $P(Z|X)= P(M|X)P(M^Z|X)$. Unlike GAN or VAE methods which map the input data into an isotropic Gaussian distribution, PM loosens the constraints on hidden probability distribution. With more difficulties for generation tasks as a price, PM enhances its effectiveness in feature extraction and disentanglement. In reality, the decoder is usually omitted in several PM applications to focus on disentangling internal representations.  Section \ref{sec:2} has suggested that directly applying unsupervised disentanglement can not bring obvious improvement. We argue that autoencoder cannot guarantee well-generalizing representation without priori knowledge. In Section \ref{sec_method},  we demonstrate how we extend adversarial PM training to debiasing tasks.

%% file: outline/method.tex
\section{Method}\label{sec_method}

\begin{figure*}[htbp]
\centerline{\includegraphics[width=0.95\textwidth]{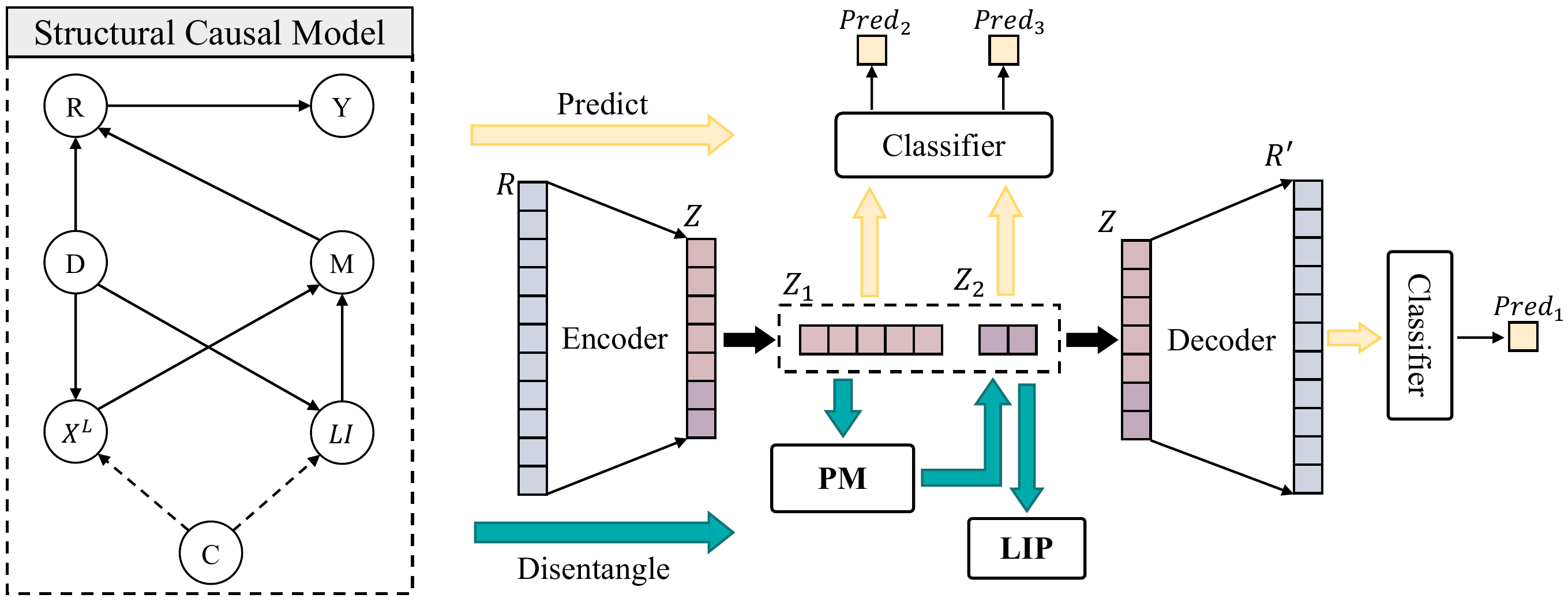}}
\caption{ Visualization of our structural causal model and CausalAPM framework. 
Under the guidance of SCM, we identify that dataset bias is originated in the  spurious correlations caused by backdoor paths. To tackle such problem, in CausalAPM, the input data is encoded into independent $Z_1$ and $Z_2$ with Disentangle process, which consists of $PM$ and $LIP $ modules. Then the disentangled representation is fed into classifier module for Predict process. Detailed explanations can be found in Section \ref{sec_method}. }
\label{fig:system}
\vspace{-0.5em}
\end{figure*}

In this section, we discuss how to train the CausalAPM model in order to learn the generalizable representation. In Section \ref{sec:4.1}, we present the SCM to formulate the causes of dataset bias and summarise our learning objectives from a causal perspective. In Section \ref{sec:4.2}, we show the model structure of CausalAPM. 
In Section \ref{sec:4.3}, we show the overall training objective.

\subsection{Structural Causal Model for NLU debiasing}\label{sec:4.1}
The left part of Figure \ref{fig:system} shows the structural causal model for NLU debiasing, containing 7 nodes in the debiasing procedure: $\boldsymbol{D}$ denotes the actual distribution of tasks corresponding to the dataset, $\boldsymbol{C}$ denotes the confounders introduced during the dataset construction, which have been observed by previous works \cite{mccoy2019right}, $\boldsymbol{X^L}$ denote the distribution of the samples with different Literal Information. $\boldsymbol{LI}$ donates the distribution of literal information. $\boldsymbol{M}$ denotes the embedding layer in the model. $\boldsymbol{R}$ denotes the representation of samples encoded by $M$, and $\boldsymbol{Y}$ denotes the labels which classifier predict.

Defining these factors, the causal process of dataset bias is observed as follows:
\begin{itemize}
\item $D \rightarrow X^L \rightarrow M$ and $D \rightarrow LI \rightarrow M$ denote the training process of the model, which constructs the training data from the real data distribution. 
\item Accompany with the construction process of the data, a backdoor path $D \rightarrow X^L \rightarrow C \rightarrow LI \rightarrow M$ is created by confounders $\boldsymbol{C}$, who introduces pseudo-correlation between the training data distribution $X^L$ and literal information LI, leading to the dataset bias. 
\end{itemize}

According to the backdoor criterion \cite{pearl1993bayesian}, we can block the path by intervention on node $LI$, which is processed with calibration formula:

\begin{small} 
\begin{align}\label{formula:2}
    P(M|do(X^L=x)) = \displaystyle\sum_{Li}P(M|X^L=x,Li)P(Li)
\end{align}
\end{small}
Where $do(X^L)$ represents intervention on variable $X^L$ to fix its value. Based on the above analysis, CausalAPM should remove the spurious correlations introduced by backdoor paths and capture the true semantic causal relations. To this end, we conduct causal interventions to debias from the literal factors. We disentangle confounder information from input representation to block the aforementioned backdoor paths in SCM, then the literal heuristic introduced by the dataset confounder will be removed. Specifically, we conduct backdoor adjustment to learn debiased NLU models, i.e., we optimize the model based on the unbiased distribution, rather than from the dataset-specific distribution.

\subsection{Causal debiasing for literal disentangle}\label{sec:4.2}

In this section, we introduce our framework for generalizable Literal Disentanglement. The right part of the Figure \ref{fig:system} exhibits our model architecture. Overall, We propose three training objectives: the basic task, APM learning, and disentangled prediction.

\subsubsection{Basic Task}
Our backbone shares similar structure with normal works to introduce NLU task-related information to trained model. Let $(X, Y)$ indicate the input data and corresponding labels.
We use Bert-base-uncased as the embedding layer to get the representation of the input data. Then we use two linear layers as encoder and decoder to get the low-dimensional representation of R, as follows:
\begin{align}
    R &= Embedding(X) \\
    Z &= Encoder(R) \\
    R^{'} &= Decoder(Z)
\end{align}

The hidden representation Z is separated into two pieces, $Z_1$ and $Z_2$, which are subsequently constrained to encode semantic and literal information respectively. In order to obtain task-relevant information, the reconstructed $R^{'}$ is imported into classifier \footnote{A single-layer FFN networks with Softmax activation} to obtain the probability of its label $Pred_1$. Based on the prediction, the basic training objective is provided:

\begin{small}
\begin{align}\label{loss:1}
    L_{base} =  -\sum_{y^i\in Y} (log(pred_1^T)y^i) +score(R,R^{i})
\end{align}
\end{small}
Where $Y$ represents the label set, $y^i$ represents an one-hot vector with 1 at the i-th position, and $score()$ represents the MSE loss function. The $Loss1$ is subsequently back-propagated to the entire Bert model, which is the one and only optimization target for Bert.

\subsubsection{APM Learning}
The analysis in Section \ref{sec:4.1} demonstrates that calibration operator on $Li$ can block the backdoor path and prompt model to fit the correct causality $P(M|do(X^L))$ with (\ref{formula:2}). To achieve this, an adversarial approach is introduced to train the disentangle encoder. Given the representation $R$ of input data, we propose two training objectives to supervise the low-dimensional representation $Z$: 1) Literal information maximization 2) Dependence minimization

\textbf{Literal information maximization} aims to extract complete literal-related information sentence representations which named informativeness \cite{cheng2020improving}. We follow \citet{eastwood2018framework} to measure the informativeness of a representation by its ability to predict the generative factor. However, previous works are supervised by predicting the bag of words of the input, which introduces extra bias to encourage the model predicting high-frequency words \cite{vasilakes2022learning}. In the $LIP$ module, we design a weaker word-independent objective, constraining the encoder to disentangle the literal information. In summary, as each piece of training data for the debiasing task consists of a pair of sentences, we use the separated representation $Z_2$ to predict sequence similarity of the sentence pairs instead of specific words. Let $X^1 = \{x_1^1,x_2^1... ,x^1_n\}$ and $X^2 = \{x_1^2,x_2^2... ,x^2_k\}$ denote the input pair, the sequence similarity $S$ and loss function is computed like the following:
\begin{align}
   & S = \frac{Card(X^1\cap X^2)}{max(n,k)} \\
   & S^{'} = LIP(Z_2) \\
   & L_{DIP} = (S^{'}-S)^2 \label{loss:2}
\end{align}
Where $Card(X)$ represents the element numbers of a collection, and $LIP(Z_2)$ represents the predicted similarity of sentence pairs.

\textbf{Dependence minimization} prevents the model from separating excessive features, which cause detriment to semantic information. In terms of disentangling, the representation of literal generating factors should lie in an independent vector space and invariant to variation on other factors \cite{higgins2018towards}. We therefore introduce the PM module shown in figure \ref{fig:system}, which acts similarly to the discriminator in GAN \cite{creswell2018generative}, aiming to predict $Z_2$ by $Z_1$ as precise as possible. The prediction acts as a supervisory signal to guide the encoder. As a result, the encoder is instructed to encode complete literal information into $Z_2$, providing an accurate representation for $LIP$ predictor and depositing residual information in $Z_1$ to maintain independence between two components. The training objectives for PM can be expressed as:
\begin{align}
    \min I(Z_1,Z_2)
\end{align}

$I(\cdot, \cdot)$ denotes the mutual information between two variables. 

Specifically, the Encoder has opposite optimization objective to the PM module, which tries to output a independent representation to keep $P(Z_2|Z_1)$ close to $P(Z_2)$. The loss function is defined as follows:
\begin{align}
   & Z_2^{'} = PM(Z_1)\\
   & L_{PM} =  \beta * Score(Z_2^{'}, Z_2) \label{loss:3}

\end{align}
Where $Score()$ represents the MSE loss function. $\beta = 1$ for training on PM module, and $\beta = -1$ for training on Encoder, respectively.

\subsubsection{Disentangled Prediction} 
The prediction are finally introduced after obtaining the disentangled representation, we complete the prediction by controlling the weight of literal information in the input. We feed $Z_1$ and $Z_2$ into the classifier respectively to obtain the probabilities of label from both semantic and literal perspectives. The two different outputs are then weighted for the final prediction. The training process can be represented as:

\begin{small}
\begin{align}
    L_{pred} =  -\sum_{y^i\in Y} (log(pred_2^T+\delta *pred_3^T)y^i) \label{loss:4}
\end{align}
\end{small}
Where $\delta$ represents the weighting parameter between semantic and literal information.

\subsection{Training} \label{sec:4.3}
Combining Eq. (\ref{loss:1}), (\ref{loss:2}), (\ref{loss:3}), and (\ref{loss:4}),  we can get the following objective function, which tries to minimize:
\begin{align}
    L =  L_{base}+\lambda * (L_{PM}+L_{DIP})+L_{pred}
\end{align}
where $\lambda$ is the temperature parameter aiming to control learning objectives for different training periods. In short, the model primarily focuses on optimizing the basic task in the early stage of training, and learning to disentangle representation afterward.

%% file: outline/exp.tex
\section{Experiments}
In this section, we verified the performance of CausalAPM on three NLU tasks and compare the results with other 9 state-of-the-art methods. We will illustrate datasets, implementation details, experimental results, and sensitivity analysis of the hyper-parameters.





\begin{table*}[htbp]
  \centering
  \caption{Model performance on MNLI, Fever, QQP, and their respective challenge test sets.}
  \begin{spacing}{1.03}
    \setlength{\tabcolsep}{1.4mm}{
    
\begin{tabular}{l|cc|ccc|ccc}
\toprule
\multicolumn{1}{c|}{\multirow{2}[4]{*}{\textbf{Model}}} & \multicolumn{2}{c|}{\textbf{MNLI}} & \multicolumn{3}{c|}{\textbf{FEVER}} & \multicolumn{3}{c}{\textbf{QQP}} \\
\cmidrule{2-9}      & ID    & HANS  & ID    & Symm. v1 & Symm. v2 & ID    & PAWS dupl  & PAWS $\neg$ dupl \\
\midrule
BERT-base & 84.3  & 61.1  & 85.4  & 55.2  & 63.1  & 91    & 96.9  & 9.8 \\
\midrule
DRiFt & 80.2  & 69.1  & 84.2  & 62.3  & 65.9  & -     & -     & - \\
Reweighting & 83.5  & 69.2  & 84.6  & 61.7  & 66.5  & 85.5  & 49.7  & 51.2 \\
Product-of-Experts & 84.1  & 66.3  & 82.3  & 62.0    & 65.9  & 88.8  & 50.3  & 61.2 \\
$\rm {{\mbox{PoE}}_{\mbox{\textbf{cross-entropy}}}}$ & 83.6  & 67.3  & 85.7  & 57.7  & 61.4  & -     & -     & - \\
$\rm {{\mbox{PoE}}_{\mbox{\textbf{self-debias}}}}$ & 80.7  & 68.5  & 85.4  & 59.7  & 65.3  & 77.4  & 44.1  & \textbf{69.4} \\
Learned-Mixin & 84.2  & 64.0    & 83.3  & 60.4  & 64.9  & 86.6  & 69.7  & 51.7 \\
Conf-reg & \textbf{84.3}  & 69.1  & 86.4  & 60.5  & 66.2  & 89.1  & 91.0    & 19.8 \\
$\rm {{\mbox{Conf-reg}}_{\mbox{\textbf{self-debias}}}}$ & 84.3  & 67.1  & 87.6  & 59.8    & 66.0  & 85.0  & 48.8  & 28.7 \\
MoCaD  & 82.3  & 70.7  & 87.1  & 65.9  & 69.1  & -     & -     & - \\
\midrule
CausalAPM(Ours)   & 84.2     & \textbf{71.1}     & \textbf{87.8}     & \textbf{66.1}     & \textbf{71.6}     & \textbf{90.6}     & 79.1     & 31.3 \\
\bottomrule
\end{tabular}%
    
     }%
    \end{spacing}
  \label{tab:r2}%
  \vspace{-1em}
\end{table*}%

\subsection{Datasets}

\begin{table}[ht] 
  \centering
  \caption{ Details of the nine state-of-the-art debiasing methods used to compare with CausalAPM  .}
  \begin{spacing}{1.0}
    \setlength{\tabcolsep}{0.8mm}{
    \begin{tabular}{l|c|c}
    \toprule
    \multicolumn{1}{c|}{\textbf{Model}} & \textbf{\makecell{requires prior \\ knowledge}} & \textbf{end-to-end} \\
    \midrule
    DRiFt  & \color[RGB]{200,85,80}\ding{52}     & \color[RGB]{200,85,80}\faTimes \\
    Reweighting & \color[RGB]{200,85,80}\ding{52}     & \color[RGB]{200,85,80}\faTimes \\
    Product-of-Experts & \color[RGB]{200,85,80}\ding{52}     & \color[RGB]{200,85,80}\faTimes \\
    $\rm {{\mbox{PoE}}_{\mbox{\textbf{cross-entropy}}}}$ & \color[RGB]{200,85,80}\ding{52}     & \color[RGB]{200,85,80}\faTimes \\
    $\rm {{\mbox{PoE}}_{\mbox{\textbf{self-debias}}}}$ & \color[RGB]{40,160,70}\faTimes     & \color[RGB]{200,85,80}\faTimes \\
    Learned-Mixin & \color[RGB]{200,85,80}\ding{52}     & \color[RGB]{200,85,80}\faTimes \\
    conf-reg & \color[RGB]{200,85,80}\ding{52}     & \color[RGB]{200,85,80}\faTimes \\
    $\rm {{\mbox{Conf-reg}}_{\mbox{\textbf{self-debias}}}}$ & \color[RGB]{40,160,70}\faTimes     & \color[RGB]{200,85,80}\faTimes \\
    MoCaD & \color[RGB]{40,160,70}\faTimes     & \color[RGB]{200,85,80}\faTimes \\
    CausalAPM(Ours) & \color[RGB]{40,160,70}\faTimes     & {\color[RGB]{40,160,70}\ding{52}} \\
    \bottomrule
    \end{tabular}%

     }
    \end{spacing}
  \label{tab:r3}%
  \vspace{-1em}
\end{table}%

The experiments are conducted on three well-known NLU tasks: natural language inference, fact verification, and paraphrase identification. The datasets used for training on each task, as well as their corresponding challenge test sets, are briefly discussed below to evaluate the impact of our debiasing methods:

\subsubsection*{Natural Language Inference}
The goal of natural language inference is to infer the relationship between the premise and the hypothesis. Recent researches \cite{mccoy-etal-2019-right,poliak-etal-2018-hypothesis} have revealed that the widely used NLI datasets contain a variety of biases. In this paper, we conduct experiments on the English Multi-Genre Natural Language Inference (MNLI) dataset \cite{N18-1101} and Heuristic Analysis for NLI Systems (HANS) \cite{mccoy-etal-2019-right}. We train the model using the training set of MNLI, and we choose MNLI-mm as the ID test set and HANS as the OOD test set.
\subsubsection*{Fact Verification}
The task is to evaluate the validity of a claim sentence in the context of a given evidence sentence, which can be categorized as support, refutes, or not enough information. We use the training dataset provided by the FEVER \cite{thorne-etal-2018-fact} for this task. Also, we use the test set of FEVER as the ID dataset and FEVER Symmetric \cite{schuster-etal-2019-towards} as the OOD dataset for evaluation.

\subsubsection*{Paraphrase Identification}
The goal of Paraphrase Identification is to identify whether a pair of statements are semantically similar. We train the model using Quora Question Pairs (QQP) dataset\footnote{\url{https://quoradata.quora.com/First-Quora-Dataset-Release-Question-Pairs}}. We perform the evaluation using QQP as ID dataset and PAWS~\cite{zhang-etal-2019-paws} as OOD dataset which consists of two types of data including \emph{duplicate} if they are paraphrased, and \emph{non-duplicate} otherwise.

\subsection{Implementation Details}
Similar to current debiasing methods, we apply our debiasing method on the uncased-bert-base model \cite{devlin-etal-2019-bert} For two sentences in a sample pair, we stitch them together and then input them into bert, and the encoding information at the [CLS] position in the output of bert will be used in the following classification task. The hyperparameters of bert are consistent with previous research papers (i.e., the learning rate is 5e-5 for MNLI and 2e-5 for FEVER and QQP, the batch size is 32 and the optimizer is AdamW with a weight decay of 0.01.).

For unique implementation in our method, we chose 64 for the hidden dimension of autoencoder, 4 for literal information and 60 for semantic information. The values of $\beta$ and $\delta$ are insensitive to specific tasks, empirically, 0.6 for $\beta$ with 0.15 for $\delta$ can achieve promised results. $\lambda$ is set to 0 for the first 2000 steps, and set to 0.6 for the rest training process.
The model is trained in an NVIDIA GeForce RTX 2080Ti GPU. All models are trained 6 epochs, and checkpoints with top-2 performance are finally evaluated on the challenge test set.



\subsection{Experimental Results and Discussions}
To fully demonstrate the generalization ability of our proposed method, we conduct experiments on three different NLU tasks and compare the results with other 9 state-of-the-art methods. The evaluation results of all methods are illustrated in table~\ref{tab:r2}. Note that previous methods \cite{mahabadi2019end,sanh2020learning,xiong2021uncertainty} have shown high variance in experiment results under different experimental settings, so we evaluate the performance of our model by randomly choosing five random seeds and report the averaged result at last.

By analyzing the experiment results of table~\ref{tab:r2} and table~\ref{tab:r3}, it is obvious that our method achieves excellent performance on the OOD dataset of all three tasks. Also, compared with other SOTA methods, our method shows the best accuracy (i.e., 71.1\%) on the HANS dataset. So, our method has the best generalization on the NLI tasks among these SOTA methods. Moreover, our method is an end-to-end approach that does not rely on any prior knowledge of the dataset (i.e., it does not require the knowledge of the type of bias existing in the dataset in advance) compared to other methods, so it achieves better usability and scalability.

It suggests that the vast majority of debiasing methods improve performance on out-of-distribution datasets by sacrificing the performance on in-distribution datasets, which means current debiasing methods attempt to achieve a trade-off between ID datasets and OOD datasets. However, our method reaches the best performance on the HANS dataset compared to all other SOTA methods, with a 10 percent improvement compared to baseline, without excessive performance degradation on ID datasets.


For the fact verification task, our method improves 10.9\% and 8.5\% relative to the baseline on the Symm. v1 and Symm. v2, respectively, which contains the best accuracy compared with other SOTA methods. Moreover, other methods are not end-to-end methods, so it has quite limited scalability, while our method can be easily expanded to other tasks. 
Our approach is designed to mitigate the damage to generalizable features while eliminating dataset bias, which is able to achieve better performance on both ID and OOD evaluation in the FEVER dataset. For the QQP dataset, our proposed method also obtains decent generalization in the PAWS dataset while minimum loss on the ID dataset. 

\subsection{Sensitivity Analysis of $\delta$ and $Z_2$}
To illustrate the effect of the involvement of literal information in decoupling and prediction on the ability to generalize, we conduct sensitivity analysis of the $\delta$ and the size of $Z_2$, i.e., the hidden dimension for literal subspace. Figure \ref{fig:system} shows the performance change under different settings of the two coefficients. The accuracy at $\delta = 0$ represents model performance with only semantic information, and accuracy at $\delta = 1$ represents model performance with full literal information. The black dotted line donates ablation results without APM objectives (with $\lambda = 0$).



we can observe that the performance is significantly enhanced with a literal rate between 0.1 and 0.3, and then decline with greater weight on literal information, while ID datasets can maintain a stable accuracy.
It proves the effectiveness of our proposed training objectives that extract and weaken literal bias while maintaining semantic information.
In addition, with only semantic information, we can still achieve significant improvements on OOD datasets, while the ID performance shows a marked decline.
This phenomenon validates the correlation between the model performance on ID datasets and the literal heuristics, which are analyzed by our SCM in Section \ref{sec:2}.

%% file: outline/relatedwork.tex
\section{Related Work}

\subsection{Disentangled Representation Learning}
Disentangled Representation Learning (DRL) aims at finding a low dimensional representation that consists of multiple explanatory and generative factors of the observational data.
\citet{bengio2013representation} and \citet{higgins2016beta} has proposed that extracted disentangled representations can improve generalization and robustness across downstream tasks. However, unsupervised disentangle methods only perform well in the simplest settings but struggle in more difficult ones \cite{zhao2018bias}.
The separated factors may consist of a combination of valid and invalid information,
which hinder its performance on debiasing tasks.

\subsection{Causal Inference for Disentanglement}
Recently, the community has raised interest in introducing causality as supervisory signals to explain disentangled latent representations, thereby improving the generalization and Interpretability of disentangling learning \cite{suter2019robustly}. \citet{kocaoglu2017causalgan} proposed CausalGAN which supports "do-operation" on images with a causal graph given as a prior.
Instead of catching independent latent factors, \citet{besserve2018counterfactuals} design a layer containing disentangled nodes representing outputs of mutually independent causal mechanisms \cite{mitrovic2020representation}.
\citet{yang2021causalvae} designed causally structured layers to disentangle factors, which enable automatically causality discovery to construct the SCM. Causal inference helps to analyze important factors of the task and provides reasonable objectives for disentangled learning.

\begin{figure}[ht]
\centering
\subfigure{
\begin{minipage}[t]{0.42\textwidth}
\centering
\includegraphics[width=\textwidth]{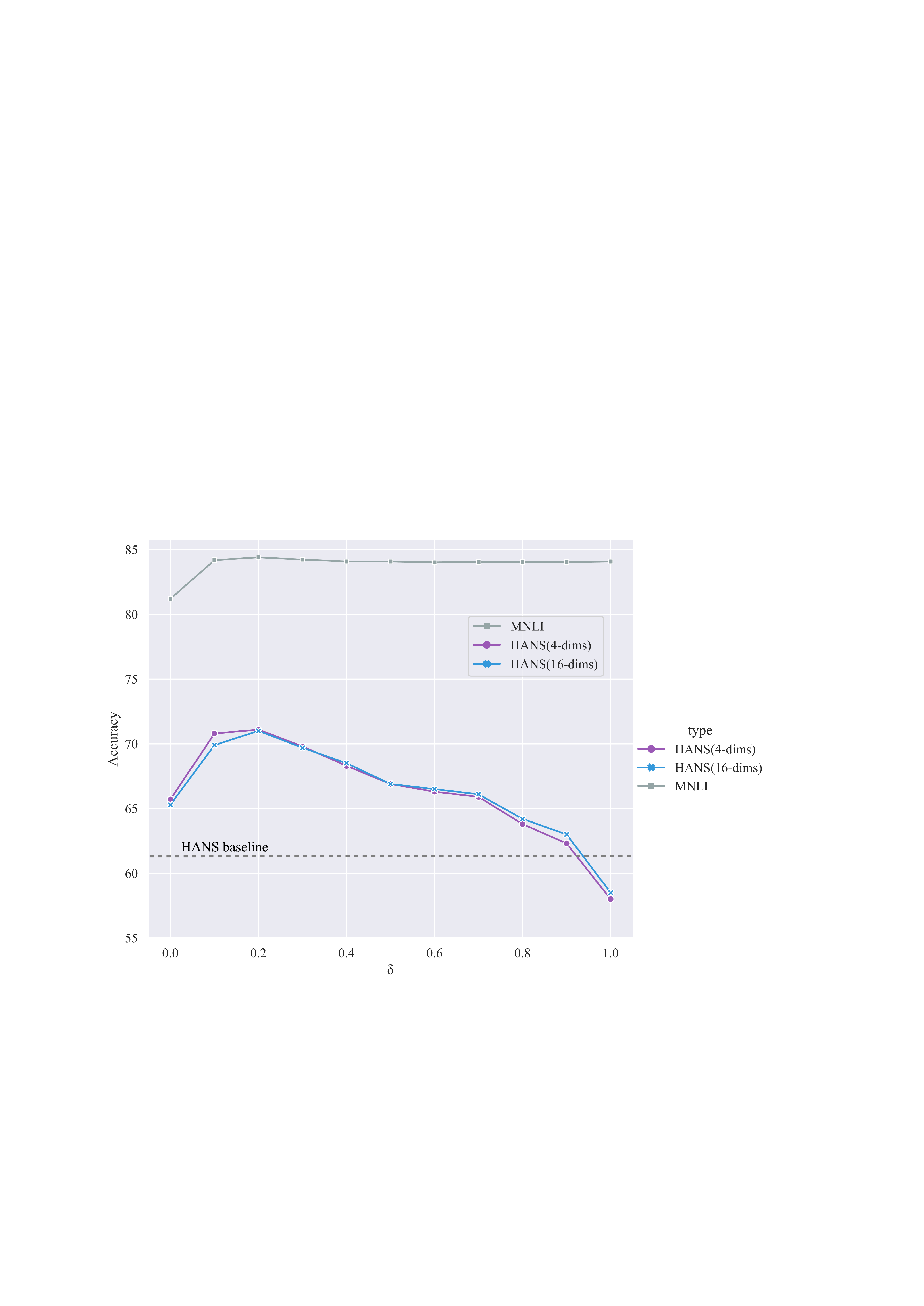}
\end{minipage}
}

\subfigure{
\begin{minipage}[t]{0.42\textwidth}
\centering
\includegraphics[width=\textwidth]{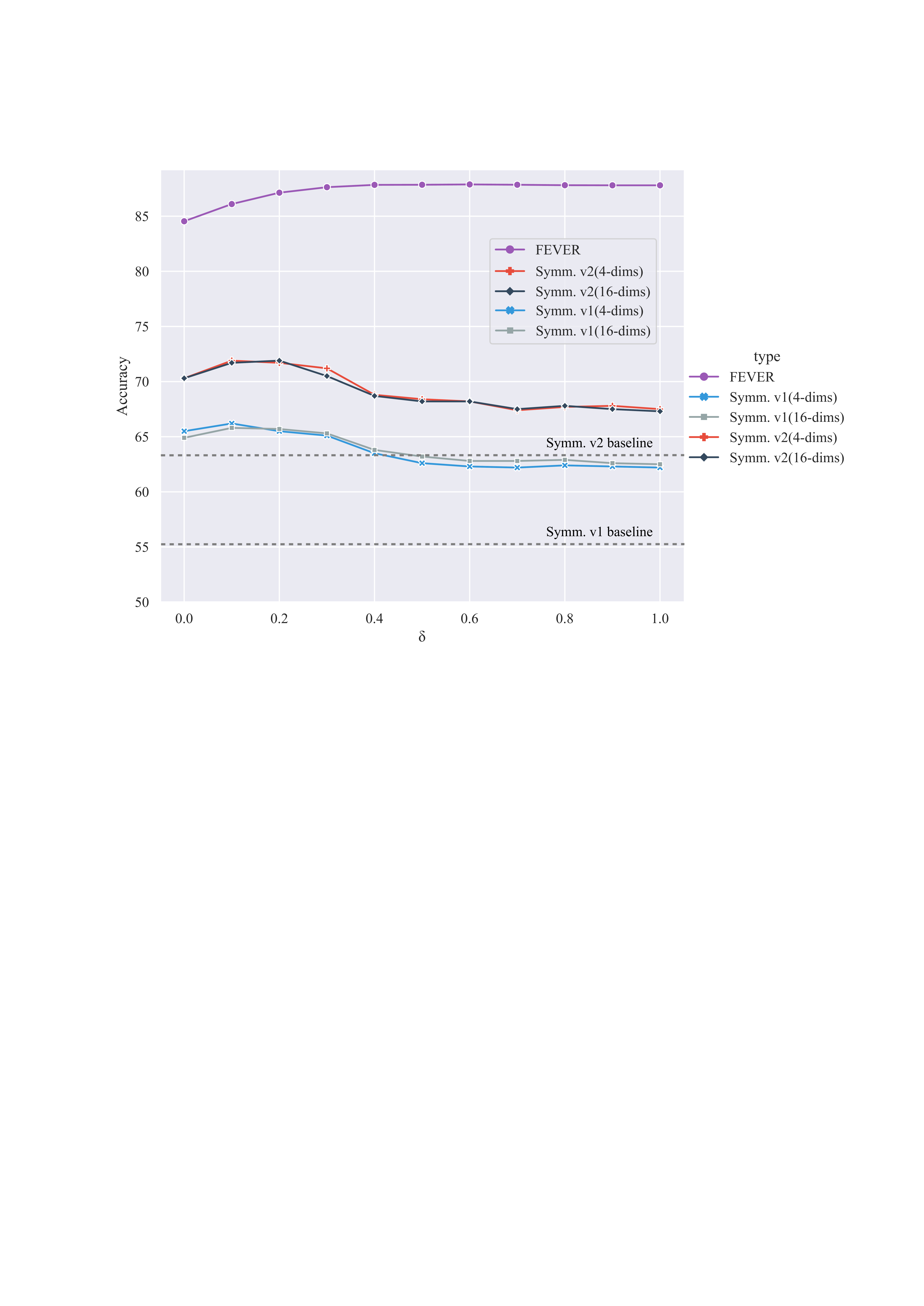}
\end{minipage}
}
\caption{ Illustrate of accuracy in different $\delta$ values. Experiments were conducted on 4 and 16 dimensional $Z_2$ with $\lambda = 0.6$, the black dotted line donates accuracy when $\lambda=0$.}
\label{fig:exp2_hans}
\vspace{-1em}
\end{figure}

%% file: outline/conclusion.tex
\section{Conclusion}

Based on the recent studies on generalization and disentanglement, we analyze how to introduce generalizable disentanglement for eliminating dataset bias. In this work, we propose a novel and flexible method - CausalAPM, to tackle the spurious correlation caused by literal heuristics. On the one hand, this framework provides a new generalizable disentangling method that separates literal and semantic information from feature granularity, on the other hand, it can effectively retain generalizable features while eliminating dataset bias. CausalAPM consists of two main learning objectives: literal information maximization, and dependence minimization. Experiments on various datasets demonstrate that CausalAPM achieves better performance on both ID and OOD datasets than comparative works.